# Research Highlights

- Propose EMD-based modeling framework with end condition methods

- Slope-based and Rato's method are recommended when selecting end condition methods

- Restraining the end effect improves the performance of EMD-based modeling framework

# Does Restraining End Effect Matter in EMD-Based Modeling Framework for Time Series Prediction? Some Experimental Evidences

Tao Xiong, Yukun Bao[*], Zhongyi Hu

Department of Management Science and Information Systems,

School of Management, Huazhong University of Science and Technology, Wuhan,

P.R.China, 430074

## Abstract

Following the "decomposition-and-ensemble" principle, the empirical mode decomposition (EMD)-based modeling framework has been widely used as a promising alternative for nonlinear and nonstationary time series modeling and prediction. The end effect, which occurs during the sifting process of EMD and is apt to distort the decomposed sub-series and hurt the modeling process followed, however, has been ignored in previous studies. Addressing the end effect issue, this study proposes to incorporate end condition methods into EMD-based decomposition and ensemble modeling framework for one- and multi-step ahead time series prediction. Four well-established end condition methods, Mirror method, Coughlin's method, Slope-based method, and Rato's method, are selected, and support vector regression (SVR) is employed as the modeling technique. For the purpose of justification and comparison, well-known NN3 competition data sets are used and four well-established prediction models are selected as benchmarks. The experimental results demonstrated that significant improvement can be achieved by the proposed EMD-based SVR models with end condition methods. The EMD-SBM-SVR model and EMD-Rato-SVR model, in particular, achieved the best prediction performances in terms of goodness of forecast measures and equality of accuracy of competing forecasts test.

---
[*] Corresponding author: Tel: +86-27-62559765; fax: +86-27-87556437.
Email: yukunbao@hust.edu.cn or y.bao@ieee.org

# 1. Introduction

Time series modeling and prediction is an area of enormous interests for both academics and practitioners. The large number of studies have compared the forecast accuracies of alternative models based on statistical theories, such as autoregressive integrated moving average (ARIMA) [1] and autoregressive conditional heteroskedasticity (ARCH) [2], or the ones based on computational intelligence, such as artificial neural networks (ANN) [3] and support vector regression (SVR) [4], for time series prediction. Existing research indicates that the latter emerges the winner, especially in short-term forecasting [5]. However, computational intelligence based forecasting models have their own shortcomings and disadvantages, e.g., local minima and over-fitting in ANN models and sensitiveness to parameter selection in both SVR and ANN models.

In view of the limitations for computational intelligence based forecasting models, recently, a hybrid empirical mode decomposition (EMD)-based modeling framework introduced by Yu et al. [6] has established itself as a promising alternative for nonlinear and nonstationary time series modeling and prediction. The attractiveness of the EMD-based modeling framework arises from the flexible decomposition-and-ensemble modeling structure resulting in a simplification of the original complicated modeling task, and the employment of EMD with which any complex signals can be decomposed into a finite number of independent and nearly periodic intrinsic mode functions (IMFs) components and a residue based purely on the local characteristic time scale [7]. As such, EMD-based modeling framework has been found to be a viable contender among various time-series models, e.g., autoregressive integrated moving average (ARIMA) [6, 8, 9], seasonal autoregressive integrated moving average (SARIMA) [10], neural networks [9, 11], and support vector regression [12], and successfully applied to different areas, including energy market [6, 11-13], tourism management [9], hydrology [14], and transportation research [10], and emergency management [8]. Regarding to the end effect occurred during the sifting process of EMD, however, the research mentioned above (see [6, 8-14]) has paid little, even no attention to, which appeals this present study.

End effect is that when calculating the upper and lower envelops with cubic spline function in the sifting process of EMD, the divergence will appear on both ends of data series, and the divergence gradually influences inside of data series so that the results are distorted badly [15], and always hurts the modeling quality as well as overall prediction performance when employing EMD-based decomposition and ensemble modeling framework for time series prediction. With regard to the problem of end effect, there has been a vast and well-established body of literatures on developing end condition methods for restraining the end effect. In general, the end condition methods are essentially to use the known points to extend both beginning and end of the series by the addition of typical waves [7, 16], extrema [17-20], or predicted values [15, 21, 22]. Although aforementioned studies have clarified the capability of these end condition methods on the restraining of end effect by means of, e.g., the orthogonality of IMFs [20], there has been very few, if any, effort to evaluate the effectiveness of end condition methods in EMD-based modeling framework for time series prediction. So, we hope this study would fill this gap.

In summary, The purpose of this study is to explicitly extend the EMD-based modeling framework with end condition methods for time series prediction, and then goes a step forward by providing the first experimental evidence within literature in which EMD-based modeling framework is applied for time series prediction to justify whether restraining the end effect is useful for achieving better prediction performance. If so, which end condition method dominates? For the purpose of justification, we conduct a large scale comparison study of EMD-based modeling framework incorporating selected end condition methods on the NN3 competition data. For the implementation of the proposed EMD-based modeling framework, support vector regression (SVR) is employed as modeling technique model in the current study in lights of that it has been found to be a viable contender among various time-series models [23-25], and successfully applied to different areas [4, 26, 27].

The paper unfolds as follows. Section 2 describes related works about EMD, end effect, and end condition methods, indicating how end effect occurs and why it is important. The details on proposed procedure of EMD with end condition methods, support vector regression, and the proposed EMD-based modeling framework incorporating selected end condition methods are presented in Section 3. Section 4 illustrates the research design on data source, preprocessing, selected counterparts, input selection, statistical criteria, methodologies implementation, and experimental procedure in details. Following that, in Section 5, the experimental results are discussed. Finally, Section 6 summarizes and concludes this work.

## 2. Related works

2.1 EMD and end effect

Empirical mode decomposition technique, first proposed by Huang et al. [7], is a kind of adaptive signal decomposition technique using the Hilbert-Huang transform (HHT) and can be applied with nonlinear and nonstationary time series. Intrinsic-mode functions (IMF) and the sifting process are the two key parts of the EMD technique. The term "intrinsic-mode function" is used because it represents the oscillation mode embedded in the data. An intrinsic-mode function is a function that satisfies two conditions: (1) in the whole data series, the number of extrema (the sum of local maxima and local minima) and the number of zero crossings must either be equal or differ at most by one; and (2) at any point, the mean value of the envelope defined by the local maxima and the envelope defined by the local minima is zero.

With this definition, IMFs can be extracted from the time series $x(t)$ according to the following sifting process:

1. Input time series $x(t)$;
2. Execute sifting process:
   (1) Initialize: $r_0(t) = x(t)$, and $i = 1$;
   (2) Extract the $i$th IMF:
      a. initialize: $h_0(t) = r_{i-1}(t), k = 1$;
      b. Identify all of the extrema (maxima and minima) of $h_{k-1}(t)$;
      c. interpolate the local maxima and local minima by a cubic spline to form upper and lower envelopes of $h_{k-1}(t)$;
      d. compute the mean $m_{k-1}(t)$ of the upper and lower envelopes just

established;

  e. create: $h_k(t) = h_{k-1}(t) - m_{k-1}(t)$;

  f. decide: if stopping criterion is satisfied then set $h_k(t) = \text{IMF}_i(t)$. Else return to step b, with $k = k+1$.

 (3) Define: $r_i(t) = r_{i-1}(t) - \text{IMF}_i(t)$;

 (4) If $r_i(t)$ is a constant or trend then sifting process can be stop, and the time series $x(t)$ is decomposed into IMFs and residue, i.e., $x(t) = \sum_{d=1}^{i} \text{IMF}_d(t) + r_i(t)$. Else return to step (2), with $i = i+1$;

3. Obtain final result, i.e., the $\text{IMF}_d(t), (1 \leq t \leq i)$, and the residue $r_i(t)$.

 As discussed in [21], however, the two ends of the time series will disperse while the series is decomposed by EMD and this disperse, termed as end effect, would "empoison" in by the whole time series gradually making the results to get distorted. To be more specific, end effect occurs during the sifting process, when the end points cannot be identified as the extrema in the procedure of 2-(2)-b above, appealing specific measure to be taken to deal with it.

 Recently, a large number of studies have developed end condition methods for restraining the end effect [7, 15-21]. Most of the proposed end condition methods are applied to "add" the extrema when end effect occurs, facilitating the construction of upper and lower envelopes during the sifting process of EMD. Details of four selected end condition methods in this study are presented in the following subsection

## 2.2 End condition methods

 In this study, we cannot examine all the end condition methods that might be useful in practice, and therefore we consider various previous literatures as guidance and focus on the most commonly used end condition methods. Finally, selected end condition methods include the Mirror method [17], Coughlin's method [16], Slope-based method [19], and Rato's method [18]. For each selected end condition methods, there are a large number of variations proposed in the literatures, and it would be a hopeless task to consider all existing varieties. Our rule is therefore to consider the basic version of each method (without the additions, or the modifications proposed by some other researchers). Detailed discussions on the selected end condition methods can be found in [16-19], but a brief introduction about their formulations is provided here. To formulate the selected end condition methods, we adopt the notations and definitions in Table 1.

<center>**<Insert Table 1 here>**</center>

1) Mirror method

For the beginning of time series $x(t)$, add local minimum $Min(0)$ by mirror symmetry with respect to the local maximum $Max(1)$; for the end of time series $x(t)$, add local maximum $Max(n+1)$ by mirror symmetry with respect to the local maximum $Min(n)$.

The newly obtained $Min(0)$ and $Max(n+1)$ are then taken for construction of the upper and lower envelopes along with initial extrema.

2) Coughlin's method

In the Coughlin's method, time series $x(t)$ is extended by additional typical waves defined as Eq. (1) instead of extrema employed in the mirror method.

$$\text{Wave extension} = A\sin\left(\frac{2\pi t}{P} + phase\right) + local\ mean. \qquad (1)$$

where the typical amplitude $A$ and period $P$ are determined by the following equations.

$$\begin{aligned} A_{begining} &= \|Max(1) - Min(1)\|, \\ A_{end} &= \|Max(n) - Min(n)\|, \\ P_{begining} &= 2\|t(Max(1)) - t(Min(1))\|, \\ P_{end} &= 2\|t(Max(n)) - t(Min(n))\|. \end{aligned} \quad (2)$$

The additional typical waves are then taken for construction of the upper and lower envelopes along with initial series so that the additional waves are continually changing in amplitude and frequency.

3) Slope-based method

For the beginning of time series $x(t)$, slope $s1$ and $s2$ are defined as Eq. (3) and Eq. (4) respectively.

$$s1 = \frac{P(2) - Q(1)}{t(Max(2)) - t(Min(1))} \quad (3)$$

$$s2 = \frac{Q(1) - P(1)}{t(Min(1)) - t(Max(1))} \quad (4)$$

Then, the time gaps between the first two successive maxima and minima are determined as $\Delta t_{max}(1) = t(Max(2)) - t(Max(1))$ and $\Delta t_{min}(1) = t(Min(2)) - t(Min(1))$. The new extrema $Min(0)$ and $Max(0)$ are updated according to the corresponding time gaps $\Delta t_{max}(1)$ and $\Delta t_{min}(1)$, and gradients $s1$ and $s2$. The ordinate and abscissa of the new extrema are positioned at

$$\begin{aligned} t(Min(0)) &= t(Min(1)) - \Delta t_{min}(1) \\ t(Max(0)) &= t(Max(1)) - \Delta t_{max}(1) \\ Q(0) &= P(1) - s1\big(t(Max(1)) - t(Min(0))\big) \\ P(0) &= Q(0) - s2\big(t(Min(0)) - t(Max(0))\big) \end{aligned} \quad (5)$$

For the end of time series, the similar procedures are used to obtain $Max(n+1)$ and $Min(n+1)$.

The newly obtained $Min(0)$, $Max(0)$, $Min(n+1)$, and $Max(n+1)$ are then taken for construction of the upper and lower envelopes along with initial extrema.

4) Rato's method

For the beginning of $x(t)$, assume $t(x(1))=0$ and $t(Max(1))>t(Min(1))$. Add local minimum $Min(0)$, where $Min(0)=Min(1)$, and $t(Min(0))=-t(Max(1))$; Add local maximum $Max(0)$, where $Max(0)=Max(1)$, and $t(Max(0))=-t(Min(1))$. For the end of time series, the similar procedures are used to obtain $Max(n+1)$ and $Min(n+1)$.

The newly obtained $Min(0)$, $Max(0)$, $Min(n+1)$, and $Max(n+1)$ are then taken for construction of the upper and lower envelopes along with initial extrema.

## 3. Methodologies

In this section, the overall formulation process of the proposed EMD-based SVR modeling framework is presented. First, EMD with end condition method is briefly introduced. Then a brief description of SVR algorithm is given. Finally, the proposed EMD-based SVR modeling framework is formulated and the corresponding steps involved are presented in details.

3.1 EMD with end condition methods

Just as mentioned in Section 2.1, the sifting process is the key part of the EMD technique and end effect occurs during the sifting process, when the end points cannot be identified as the extrema, appealing end condition methods to be incorporated into the sifting process. The improved sifting process with end condition method is depicted in Fig.1.

<Insert Fig.1 here>

3.2 Support vector regression

Support vector regression (SVR), first proposed by Vapnik et al. [28] based on the structured risk minimization principle, is found to be a viable contender among various time series models [4, 29] by minimizing an upper bound of the generalization error. Here, SVR is used as tool for forecasting. This subsection gives a brief description of SVR. The details of the formulation can be found in [28].

Given a set of data $\{x_t, y_t\}(t=1,2…,T)$, where $x_t \in \Re^T$ is the $t$th input pattern and $y_t$ is its corresponding observed result, the basic idea of SVR is first to map the original data $x_t$ into a high-dimensional feature space via a nonlinear mapping function $\varphi(\cdot)$, then to make linear regression in this high-dimension feature space and find the optimal separating hyperplane with minimal classification errors [12].

In general, SVR approximate the function using the following:

$$f(x) = w^T \varphi(x) + b \qquad (6)$$

where $\varphi(x)$ is the nonlinear function mapping from input space $x$ into a high-dimensional feature space, and $f(x)$ is the estimated value. Coefficients $w^T$ and $b$ are obtained by minimizing the regularized risk function, which can be transformed into the following optimization problem:

$$\begin{cases} \min & \dfrac{1}{2}w^T w + \gamma \sum_{t=1}^{T}\left(\zeta_t + \zeta_t^*\right) \\ s.t. & w^T \varphi(x_t) + b - y_t \le \varepsilon + \zeta_t^*,\ (i=1,2,\ldots,T) \\ & y_t - \left(w^T \varphi(x_t) + b\right) \le \varepsilon + \zeta_t,\ (i=1,2,\ldots,T) \end{cases} \qquad (7)$$

where $\gamma$ is the penalty parameter, and nonnegative variables $\zeta_t$ and $\zeta_t^*$ are the slack variables which represent the distances from actual value to the corresponding boundary value of $\varepsilon-$tube.

So the problem of constructing the optimal hyperplane is transformed into the following the quadratic programming problem:

$$\begin{cases} \min\limits_{a,a^*} & \dfrac{1}{2}\sum_{i=1}^{n}\sum_{j=1}^{n}(a_i - a_i^*)(a_j - a_j^*)(\phi(x_i)\cdot\phi(x_j)) + \sum_{i=1}^{n}(\varepsilon - y_i)a_i + \sum_{i=1}^{n}(\varepsilon + y_i)a_i^* \\ s.t. & \sum_{i}^{n}(a_i - a_i^*) = 0,\ a_i, a_i^* \in [0,C] \end{cases} \qquad (8)$$

where $a_i$ and $a_i^*$ are corresponding Lagrange multipliers used to push and pull $f(x_i)$ towards the outcome of $y_i$ respectively.

The decision function can be shown as:

$$f(x) = \sum_{i=1}^{n}(a_i - a_i^*)K(\mathbf{x}_i, \mathbf{x}_j) + b. \qquad (9)$$

$K(\mathbf{x}_i, \mathbf{x}_j) = (\phi(x_i)\cdot\phi(x_j))$ is defined as the kernel function. The elegance of using the kernel function is that one can deal with feature spaces of arbitrary dimensionality without having to compute the map $\phi(x)$ explicitly. In this study, we select a common kernel function, i.e., RBF function, $K(\mathbf{x}_i, \mathbf{x}_j) = \exp\left(-\gamma\|\mathbf{x}_i - \mathbf{x}_j\|^2\right), \gamma > 0$, as the kernel function.

### 3.3 The proposed prediction models

It should be noted that SVR is employed as modeling technique in this study. As such, this study turns out to develop different prediction models under EMD-based modeling framework with or without end condition methods using SVR, i.e., EMD-based SVR modeling framework for short.

As shown in Fig. 2, the proposed EMD-based SVR modeling framework is generally composed of the following three main steps:

*Step* 1: The original series are first decomposed into a finite and often a small number of intrinsic mode functions (IMFs) and a residue using EMD technique. In the sifting process of EMD, selected end condition method is applied to restrain the end effect following the procedures illustrated in the above subsection.

*Step* 2: After the components (IMFs and a residue) are adaptively extracted via EMD, each component is modeled by an independent SVR model to forecast the component series respectively.

*Step* 3: The forecasts of all components are aggregated using another independent SVR model, which model the relationship among the IMFs and the residue, to produce an ensemble forecasts for the original series.

<center>**<Insert Fig.2 here>**</center>

Following the EMD-based SVR modeling framework, different prediction models can be developed. For example, in case mirror method is selected and incorporated into the EMD to deal with end effect, then EMD-MM-SVR prediction model is derived. Following the same naming rule, EMD-Coughlin-SVR, EMD-SBM-SVR, and EMD-Rato-SVR refer to the prediction models with corresponding end condition methods respectively. It should be noted that EMD-SVR refers to the model without any end condition methods.

## 4. Research design

This section provides details about the research design. In section 4.1, the details of the data sets and relating data preprocessing procedure are given. Section 4.2 presents the selected counterparts for comparison. The input selection is briefly presented in Section 4.3. Section 4.4 lists and briefly describes the goodness of forecast measures and equality of accuracy of competing forecasts test used. Section 4.5 presents the implementations of EMD, Wavelet, SVR, and SARIMA. Section 4.6 depicts the experimental procedures with NN3 competition data in details.

### 4.1 The datasets and data preprocessing

The datasets of 111 time series distributed for the NN3 competition are used for this study[1]. This competition was organized in 2007, and targeted at computational intelligence based forecasting approaches. The data are monthly, with positive observations and structural characteristics which vary widely across the time series. Many of the series are dominated by a strong seasonal structure (e.g. #55, #57 and #73), while some series exhibit both trending and seasonal behavior (e.g. #1, #11 and #12). We leave the last 18 months of observations for evaluating and comparing the out-of-sample prediction performances of the proposed models against selected counterparts. All performance comparisons are based on these $18 \times 111$ out of sample points.

Since most of the time series considered exhibit strong seasonal component or trend pattern as shown in Fig.3, we conduct deseasonalizing by means of the revised multiplicative seasonal decomposition presented in [30]. In addition, detrending is performed by fitting a polynomial time trend and then subtracting the estimated trend from the series when trends are detected by the Mann-Kendall test [31].

---

[1] The datasets can be obtained from http://www.neural-forecasting-competition.com/NN3/datasets.htm

<Insert Fig.3 here>

## 4.2 The selected counterparts for comparison

Single SVR, Seasonal ARIMA (SARIMA) and Wavelet-SVR [32] are selected as counterparts for the purpose of comparison. It should be noted the reason for selecting single SVR is to justify the effectiveness of EMD-based modeling framework, for the selection of SARIMA is due to the exhibited characteristics of strong seasonality of the NN3 data sets, and for the selection of Wavelet-SVR is the similar modeling mechanism shared by EMD-based and Wavelet-based modeling frameworks (However, the present study focuses on the technical improvement on EMD-based modeling framework addressing the issue of end effect, but not the comparative study between EMD and Wavelet though it could be an interesting topic worthy of further exploration). The essential formulations of SARIMA and Wavelet-SVR have been presented in many papers, so will not be repeated here to keep this paper concise. For detailed introduction to these methods, please refer to [10, 32].

Additionally, the performances on both one-step-ahead (prediction horizon $H=1$) and multi-step-ahead (prediction horizon $H=18$) prediction are compared across all the models to provide more evidences for justification. Note that the iterated strategy for multi-step-ahead prediction is employed in this study due to its simplicity and popularity in literature [11, 33]. This strategy constructs a prediction model by means of minimizing the squares of the in-sample one-step-ahead residuals, and then uses the predicted value as an input for the same model when we forecast the subsequent point, and continue in this manner until reaching the horizon.

4.3 Input selection

Filter method is employed for input selection in this study. In the case of the filter method, the best subset of inputs is selected a priori based only on the dataset. The input subset is chosen by an evaluation criterion, which measures the relationship of each subset of input variables with the output [34]. Specifically, in terms of evaluation criteria, the partial mutual information[2] [35] is used for the prediction models. Mutual information (MI) is a commonly adopted measure of dependence between variables and has been widely used for input selection [34]. However, this raises a major redundancy issue redundancy issue because the MI criterion does not account for the interdependency between candidate variable. To address this issue, Sharma [35] developed an improved algorithm that exploits the concept of partial mutual information (PMI), which is the nonlinear statistical analog of partial correlation. The definitions of PMI are shown as follows:

$$PMI = \iint f_{X',Y'}(x',y') \ln\left[\frac{f_{X',Y'}(x,y)}{f_{X'}(x) f_{Y'}(y)}\right] dx' dy' \qquad (10)$$

$$x' = x - E[x|\mathrm{z}] \qquad (11)$$

---

[2] The Matlab code can be obtained from
http://www.cs.tut.fi/~timhome/tim-1.0.2/tim/matlab/mutual_information_p.m.htm

$$y' = y - E[y|z] \tag{12}$$

where $X'$ and $Y'$ are generalized to represent time series $x(t)$ and lagged time $x(t-i)$ with time step $i$ $(i \leq d)$ conditional on $Z$ which is a set of remaining time-lag variables. In performing the PMI, the input variable that has the highest conditional PMI value at each iteration is added to the selection set. The maximum embedding order $d$ is set to 12 for the input selection process over all the series from NN3 competition data sets [36].

## 4.4 Statistical criteria

It should be noted that the impact of end condition methods on the quality of EMD has been widely investigated in [7, 15-22] and it is not the focus of the current study, but the impact of end condition methods on prediction performance of EMD-based modeling framework for time series prediction has not been widely explored which is the research goal of this study. Hence, statistical criteria such as goodness of forecast measures (i.e., symmetric mean absolute percentage error (SMAPE) and mean absolute scaled error (MASE)) and equality of accuracy of competing forecasts test (i.e., one-way analysis of variance (ANOVA) and Tukey honestly significant difference (HSD) test) are employed here.

To compare the effectiveness of the different prediction models, no single accuracy measure can capture all the distributional features of the errors when summarized across data series. Here, we consider two forecast accuracy measures. The first is the SMAPE defined as Eq. (13), as this is the main measure considered in NN3 competition [37]. The second accuracy measure is the MASE, defined as Eq.

(14). It has recently been suggested by Hyndman and Koehler [38] as a means of overcoming observation and errors around zero existing in some measures. The MASE has some features which are better than the SMAPE, which has been criticized for the fact that its treatment of positive and negative errors is not symmetric [39]. However, because of its widespread use, the SMAPE will still be used in this study. The smaller the values of SMAPE and MASE, the closer are the predicted time series values to the actual values.

$$\text{SMAPE} = \frac{1}{M \cdot T} \sum_{m=1}^{M} \sum_{t=1}^{T} \left| \frac{x_m(t) - \hat{x}_m(t)}{(|x_m(t)| + |\hat{x}_m(t)|)/2} \right| \tag{13}$$

$$\text{MASE} = \frac{1}{M \cdot T} \sum_{m=1}^{M} \sum_{t=1}^{T} \left( \left| \frac{x_m(t) - \hat{x}_m(t)}{\frac{1}{N-1} \sum_{i=2}^{N} |x_m(i) - x_m(i-1)|} \right| \right) \tag{14}$$

where $x_m(t)$ denotes the observation at period $t$ for time series $m$, $\hat{x}_m(t)$ denotes the forecast of $x_m(t)$, $M$ is the number of time series (in this case, $M = 111$), $T$ is the number of observation in the hold-out sample (in this case, $T = 18$), and $N$ is the number of observation in the estimation sample for time series $m$.

In this study, we repeat running each model fifty times for NN3 dataset to even out the fluctuations. Then each of the fifty runs will produce a SMAPE for all 111 time series. Next, the mean and standard deviation of these fifty SMAPE are calculated and listed in the tables for examining the performance of different models. Similarly, the mean and standard deviation of MASE are also computed. Note that the error measures are computed after rolling back of the preprocessing step performed, such as deseasonalization and detrending.

Following [40], we also conduct a number of statistical tests to compare each model based on the obtained fifty SMAPE and MASE at the 0.05 significance level. For each prediction horizon ($H = 1$ and $18$) and performance measure (i.e., SMAPE and MASE), we perform a one-way analysis of variance (ANOVA) procedure to determine if there exists statistically significant difference among the eight models in out-of-sample forecasting. Then, to further identify the significant difference between any two models, the Tukey honestly significant difference (HSD) test [41] is used to compare all pairwise differences simultaneously. Note that Tukey HSD test is a post-hoc test, this means that a researcher should not perform Tukey HSD test unless the results of ANOVA are positive.

## 4.5 Methodologies implementations

In this study, EMD[3] is implemented using the program provided by Huang et al. [7]. The number of sifting passes for IMF extraction is fixed at 10, and the whole sifting process stops after $log_2 N$ IMFs have been extracted, where $N$ is the length of the data series.

The Wavelet toolbox in Matlab is used to implement the discrete Wavelet transform. This step involves several different families of Wavelets and a detailed comparison of their performance. In this study, the Daubechies's Wavelets of order 7 is adopted through preliminary simulation in a trial-error fashion. To determine the number of decomposition levels, $L = \text{int}\left[\log(N)\right]$ is used [42]. L presents the decomposition level while $N$ denotes the length of the data series.

---

[3] Matlab code are available at http://rcada.ncu.edu.tw/

LibSVM (version 2.86)[4] [43] is employed for SVR modeling here. We select the Radial basis function (RBF) as the kernel function in the EMD-based prediction models when modeling the IMFs data. The linear kernel function is selected to model the relationship among the IMFs and the residue due to its simplicity and better performances after extensive experimental trials on different kernel functions. To determine the hyper-parameters, namely $C, \varepsilon, \gamma$ (in the case of RBF as the kernel function), a population-based search algorithm, named particle swarm optimization (PSO) [44], is employed in the current study. Due to its simplicity and generality as no important modification was made for applying it to model selection, PSO has been recently established for parameter determination of SVR [45]. In solving hyper-parameter selection by the PSO, each particle is requested to represent a potential solution $(C, \varepsilon, \gamma)$. Concerning the selection of parameters (i.e., cognitive and interaction coefficients, swarm size, and number of iterations) in binary PSO, it is yet another challenging model selection task. Fortunately, several empirical and theoretical studies have been performed about the parameters of PSO from which valuable information can be obtained [46]. In this study, the parameters are determined according to the recommendations in these studies and selected based on the prediction performance and computational time in a trial-error fashion. Through experiment, the parameter values of PSO are set as follows. Both the cognitive and interaction coefficients are set to 2. The swarm size and number of iterations are set to be 10 and 50, respectively.

---

[4] Matlab code are available at http://www.csie.ntu.edu.tw/~cjlin/libsvm/

For SARIMA estimation, the automatic model selection algorithm proposed by Hyndman and Khandakar [47] and implemented in the R software package 'forecast'[5] is used in this study.

It should be noted that in the model estimation stage for EMD- and Wavelet-based SVR models, all the samples from training sets are decomposed at one time and used for model estimation. 10 fold cross validation is used for parameters tuning under the commonly used grid search. Finally, the achieved model based on training sets is tested on hold-out sample in the way as the decomposition is repeated with a next data added.

## 4.6 Experimental procedure

Fig. 4 shows the procedure for performing experiments with the NN3 competition data in this study. Each series is split into the estimation sample and the hold-out sample firstly. Then, the optimal eight examined models for estimation sample is determined. Afterwards, obtained eight models are used for one- and multi-step-ahead time series prediction for hold-out sample and the two accuracy measures are computed. Furthermore, the modeling process for each series is repeated fifty times. Upon the termination of this loop, performance of the examined models is judged in terms of the mean and standard deviation of the SMAPE and MASE of fifty replications. In addition, the ANOVA and Tukey HSD tests are used to test the statistical significance of any two competing prediction models at the 0.05 significance level.

---

[5] R package 'forecast' are available at http://ftp.ctex.org/mirrors/CRAN/

<Insert Fig.4 here>

## 5. Results and discussions

The prediction performances of all the examined models (i.e., EMD-Rato-SVR, EMD-Coughlin-SVR, EMD-SBM-SVR, EMD-MM-SVR, EMD-SVR, Wavelet-SVR, SVR, and SARIMA) in terms of mean and standard deviation of two accuracy measures (i.e., SMAPE and MASE) for one- and multi-step-ahead prediction are shown in Table 2. As per the results presented, one can deduce the following observation:

<Insert Table 2 here>

- Overall, the proposed prediction models (these are, EMD-MM-SVR, EMD-Coughlin-SVR, EMD-SBM-SVR, and EMD-Rato-SVR) outperform the EMD-based SVR prediction model without any end condition methods (that is, EMD-SVR) without exception. As such, we argue that the superior performance of proposed prediction models relative to EMD-SVR as a result of restraining the end effect occurred during the sifting process of EMD.

- The proposed EMD-SBM-SVR and EMD-Rato-SVR outperform the EMD-MM-SVR and EMD-Coughlin-SVR regardless of the accuracy measures and prediction horizon considered, indicating the superiority of Slope-based method and Rato's method as end condition methods in EMD-based modeling framework from the perspective of time series prediction.

- The six hybrid ensemble models (i.e., EMD-MM-SVR, EMD-Coughlin-SVR, EMD-SBM-SVR, EMD-Rato-SVR, EMD-SVR, and Wavelet-SVR) consistently

achieve more accurate forecasts than the two single models (i.e., SVR and SARIMA). The main reason could be that the decomposition strategy does effectively improve prediction performance.

- As far as the comparison between the EMD-SVR and Wavelet-SVR, they are almost a tie and the results are mixing among the prediction measures and horizons examined. In terms of SMAPE, EMD-SVR wins for one-step-ahead prediction but loses for eighteen-step-ahead prediction. In terms of MASE, EMD-SVR loses for one-step-ahead prediction but wins for eighteen-step-ahead prediction.

- When comparing single prediction models, the SARIMA model mostly ranks the last, while the SVR can produce far better results. The possible reason is that SARIMA is a typical linear model not suitable for capturing nonlinear patterns hiding in the NN3 dataset.

For each performance measure and prediction horizon, we perform an ANOVA procedure to determine if there exists statistically significant difference among the eight models in hold-out sample. Table 3 shows the results of ANOVA test, from which we can see that the all the ANOVA results are significant at the 0.05 significance level, suggesting that there are significant differences among the eight models. To further identify the significant difference between any two models, the Tukey's HSD test is used to compare all pairwise differences simultaneously here. Table 4 shows the results of these multiple comparison tests at 0.05 significance level (for abbreviation, we use SBM, Rato, Coughlin, and MM in replace of EMD-SBM-SVR, EMD-Rato-SVR, EMD-Coughlin-SVR, and EMD-MM-SVR respectively in this table). For each accuracy measure and prediction horizon, we rank order the models from 1 (the best) to 8 (the worst). Several observations can be made from Table 4.

- When considering one-step-ahead prediction, EMD-SBM-SVR and EMD-Rato-SVR significantly outperform the EMD-SVR across two measures.

- However, when considering multi-step-ahead prediction, all the proposed four prediction models significantly outperform the EMD-SVR across two measures.

- The EMD-SBM-SVR and EMD-Rato-SVR significantly outperform the EMD-Coughlin-SVR and EMD-MM-SVR, with one exception, where EMD-MM-SVR performs the poorest at 95% statistical confidence level.

- There is no significant difference of prediction performance between EMD-SBM-SVR and EMD-Rato-SVR. One exception occurs when $H = 1$ and SMAPE is used, in which the EMD-SBM-SVR significantly outperform the EMD-Rato-SVR.

- As far as the comparison EMD-SVR vs. Wavelet-SVR is concerned, the difference in prediction performance is not significant at the 0.05 level in all cases.

- For each accuracy measure and prediction horizon, the hybrid ensemble models significantly outperform the single models.

- When comparing single prediction models, the SVR performs significantly better than SARIMA without exception.

- The SARIMA performs the poorest at 95% statistical confidence level in all cases.

## 6. Conclusions

This study contributed to propose an extension to well-established EMD-based modeling framework by incorporating end condition methods for time series prediction, and provide large scale experimental evidences for the purpose of justification. The experimental results lead to the following main conclusions. (1) The original EMD-based modeling framework is outperformed by the proposed four variants with different end condition methods, confirming the helpfulness of restraining the end effect in the context of time series modeling and prediction. (2) EMD-SBM-SVR and EMD-Rato-SVR achieved better as well as more stable prediction performances than the other counterparts in terms of rank-based measure, indicating the superiority of slope-based method and Rato's method as end condition methods.

The limitations of this study lie in two aspects. First, although we have examined a variety of end condition methods that are most commonly used in EMD literatures, there are many other possible methods in restraining the end effect of EMD, which may shed a different light on the modeling issue. Second, Furthermore, EEMD, recently proposed by Wu and Huang [48], is a substantial improvement over the original EMD, which may shed a different light on the modeling issue and further study to this regard is solicited.

## Acknowledgement

The authors would like to thank the anonymous reviewers for their valuable suggestions and constructive comments. This work was supported by the Natural Science Foundation of China (Grant No. 70771042) and the Fundamental Research Funds for the Central Universities (2012QN208-HUST) and a grant from the Modern Information Management Research Center at Huazhong University of Science and Technology.

# Reference

[1] Y.S. Lee, L.I. Tong, Forecasting time series using a methodology based on autoregressive integrated moving average and genetic programming. Knowl-based Syst 24 (2011) 66-72.

[2] C.W. Cheong, Modeling and forecasting crude oil markets using ARCH-type models. Energ Policy 37 (2009) 2346-2355.

[3] C.M. Lee, C.N. Ko, Time series prediction using RBF neural networks with a nonlinear time-varying evolution PSO algorithm. Neurocomputing 73 (2009) 449-460.

[4] K. Kim, Financial time series forecasting using support vector machines. Neurocomputing 55 (2003) 307-319.

[5] T. Hill, M. O'Connor, W. Remus, Neural Network Models for Time Series Forecasts. Manage Sci 42 (1996) 1082-1092.

[6] L.A. Yu, S.Y. Wang, K.K. Lai, Forecasting crude oil price with an EMD-based neural network ensemble learning paradigm. Energy Econ 30 (2008) 2623-2635.

[7] N.E. Huang, Z. Shen, S.R. Long, M.C. Wu, H.H. Shih, Q. Zheng, et al., The empirical mode decomposition and the Hilbert spectrum for nonlinear and non-stationary time series analysis. Proceedings of the Royal Society of London Series A: Mathematical, Physical and Engineering Sciences 454 (1998) 903-995.

[8] X. Xu, Y. Qi, Z. Hua, Forecasting demand of commodities after natural disasters. Expert Syst Appl 37 (2010) 4313-4317.

[9] C.F. Chen, M.C. Lai, C.C. Yeh, Forecasting tourism demand based on empirical mode decomposition and neural network. Knowl-based Syst 26 (2012) 281-287.

[10] Y. Wei, M.C. Chen, Forecasting the short-term metro passenger flow with empirical mode decomposition and neural networks. Transportation Research Part C: Emerging Technologies 21 (2012) 148-162.

[11] Z. Guo, W. Zhao, H. Lu, J. Wang, Multi-step forecasting for wind speed using a modified EMD-based artificial neural network model. Renew Energ 37 (2012) 241-249.

[12] L. Tang, L. Yu, S. Wang, J. Li, S. Wang, A novel hybrid ensemble learning paradigm for nuclear energy consumption forecasting. ApEn 93 (2012) 432-443.

[13] Y. Dong, J. Wang, H. Jiang, J. Wu, Short-term electricity price forecast based on the improved hybrid model. Energ Convers Manage 52 (2011) 2987-2995.

[14] G. Napolitano, F. Serinaldi, L. See, Impact of EMD decomposition and random initialisation of weights in ANN hindcasting of daily stream flow series: An empirical examination. J Hydrol 406 (2011) 199-214.

[15] Y. Deng, W. Wang, C. Qian, Z. Wang, D. Dai, Boundary-processing-technique in EMD method and Hilbert transform. Chin Sci Bull 46 (2001) 954-960.

[16] K. Coughlin, K.K. Tung, 11-year solar cycle in the stratosphere extracted by the empirical mode decomposition method. Adv Space Res 34 (2004) 323-329.

[17] G. Rilling, P. Flandrin, P. Gonçalvés, On empirical mode decomposition and its algorithms. IEEE-EURASIP Workshop on Nonlinear signal and Image Processing(2003) 1-5.

[18] R. Rato, M. Ortigueira, A. Batista, On the HHT, its problems, and some solutions. Mech Syst Signal Pr 22 (2008) 1374-1394.

[19] M. Dätig, T. Schlurmann, Performance and limitations of the Hilbert–Huang transformation (HHT) with an application to irregular water waves. OcEng 31 (2004) 1783-1834.

[20] F. Wu, L. Qu, An improved method for restraining the end effect in empirical mode decomposition and its applications to the fault diagnosis of large rotating machinery. J Sound Vib 314 (2008) 586-602.

[21] J. Cheng, D. Yu, Y. Yang, Application of support vector regression machines to the processing of end effects of Hilbert–Huang transform. Mech Syst Signal Pr 21 (2007) 1197-1211.

[22] D.C. Lin, Z.L. Guo, F.P. An, F.L. Zeng, Elimination of end effects in empirical mode decomposition by mirror image coupled with support vector regression. Mech Syst Signal Pr 31 (2012) 13–28.

[23] P.F. Pai, W.C. Hong, Support vector machines with simulated annealing algorithms in electricity load forecasting. Energ Convers Manage 46 (2005) 2669-2688.
[24] H. Prem, N.R.S. Raghavan, A support vector machine based approach for forecasting of network weather services. Journal of Grid Computing 4 (2006) 89-114.
[25] K.Y. Chen, C.H. Wang, Support vector regression with genetic algorithms in forecasting tourism demand. Tourism Management 28 (2007) 215-226.
[26] P.S. Yu, S.T. Chen, I.F. Chang, Support vector regression for real-time flood stage forecasting. J Hydrol 328 (2006) 704-716.
[27] D. Niu, D. Liu, D.D. Wu, A soft computing system for day-ahead electricity price forecasting. Applied Soft Computing 10 (2010) 868-875.
[28] V. Vapnik, S.E. Golowich, A. Smola, Support vector method for function approximation, regression estimation, and signal processing. Advances in neural information processing systems (1997) 281-287.
[29] Q. Li, Q. Meng, J. Cai, H. Yoshino, A. Mochida, Applying support vector machine to predict hourly cooling load in the building. ApEn 86 (2009) 2249-2256.
[30] R.R. Andrawis, A.F. Atiya, H. El-Shishiny, Forecast combinations of computational intelligence and linear models for the NN5 time series forecasting competition. Int J Forecasting 27 (2011) 672-688.
[31] B. Onoz, M. Bayazit, The power of statistical tests for trend detection. Turkish Journal of Engineering and Environmental Sciences 27 (2003) 247-251.
[32] O. Kisi, M. Cimen, Precipitation forecasting by using wavelet-support vector machine conjunction model. Eng Appl Artif Intel 25 (2012) 783-792.
[33] L. Tang, L. Yu, S. Wang, J. Li, S. Wang, A novel hybrid ensemble learning paradigm for nuclear energy consumption forecasting. ApEn 93 (2011) 432-443.
[34] A. Sorjamaa, J. Hao, N. Reyhani, Y. Ji, A. Lendasse, Methodology for long-term prediction of time series. Neurocomputing 70 (2007) 2861-2869.
[35] A. Sharma, Seasonal to interannual rainfall probabilistic forecasts for improved water supply management: Part 1—A strategy for system predictor identification. J Hydrol 239 (2000) 232-239.
[36] S. Ben Taieb, A. Sorjamaa, G. Bontempi, Multiple-output modeling for multi-step-ahead time series forecasting. Neurocomputing 73 (2010) 1950-1957.
[37] S.F. Crone, K. Nikolopoulos, M. Hibon, Automatic Modelling and Forecasting with Artificial Neural Networks–A forecasting competition evaluation. IIF/SAS Grant 2005 Research Report 2008).
[38] R.J. Hyndman, A.B. Koehler, Another look at measures of forecast accuracy. Int J Forecasting 22 (2006) 679-688.
[39] P. Goodwin, R. Lawton, On the asymmetry of the symmetric MAPE. Int J Forecasting 15 (1999) 405-408.
[40] M. Qi, G.P. Zhang, Trend time–series modeling and forecasting with neural networks. Neural Networks, IEEE Transactions on 19 (2008) 808-816.
[41] B. Tukey's, Multiple comparisons.　(1953).
[42] V. Nourani, M.T. Alami, M.H. Aminfar, A combined neural-wavelet model for prediction of Ligvanchai watershed precipitation. Eng Appl Artif Intel 22 (2009) 466-472.
[43] C.C. Chang, C.J. Lin, LIBSVM: a library for support vector machines. ACM Transactions on Intelligent Systems and Technology (TIST) 2 (2011) 27.
[44] R.C. Eberhart, Y. Shi, J. Kennedy. Swarm intelligence: Elsevier; 2001.
[45] X. Guo, J. Yang, C. Wu, C. Wang, Y. Liang, A novel LS-SVMs hyper-parameter selection based on particle swarm optimization. Neurocomputing 71 (2008) 3211-3215.
[46] J.F. Kennedy, J. Kennedy, R.C. Eberhart. Swarm intelligence: Morgan Kaufmann Pub; 2001.
[47] R.J. Hyndman, Y. Khandakar, Automatic Time Series for Forecasting: The Forecast Package for R. journal of statistical software 26 (2007) 1-22.
[48] Z. Wu, N.E. Huang, Ensemble empirical mode decomposition: A noise-assisted data analysis method. Advances in Adaptive Data Analysis 1 (2009) 1-41.

# Caption page

Table 1: Notation and definition for end condition methods

Table 2: Prediction accuracy measure of different models for hold-out sample

Table 3: ANOVA results for hold-out ample

Table 4: Tukey HSD test results with ranked models for hold-out sample

Fig. 1: The EMD with end condition method

Fig. 2: The EMD-based SVR modeling framework

Fig. 3: Two representative NN3 time series (#12 and #55)

Fig. 4: Experiment procedures

# Figures

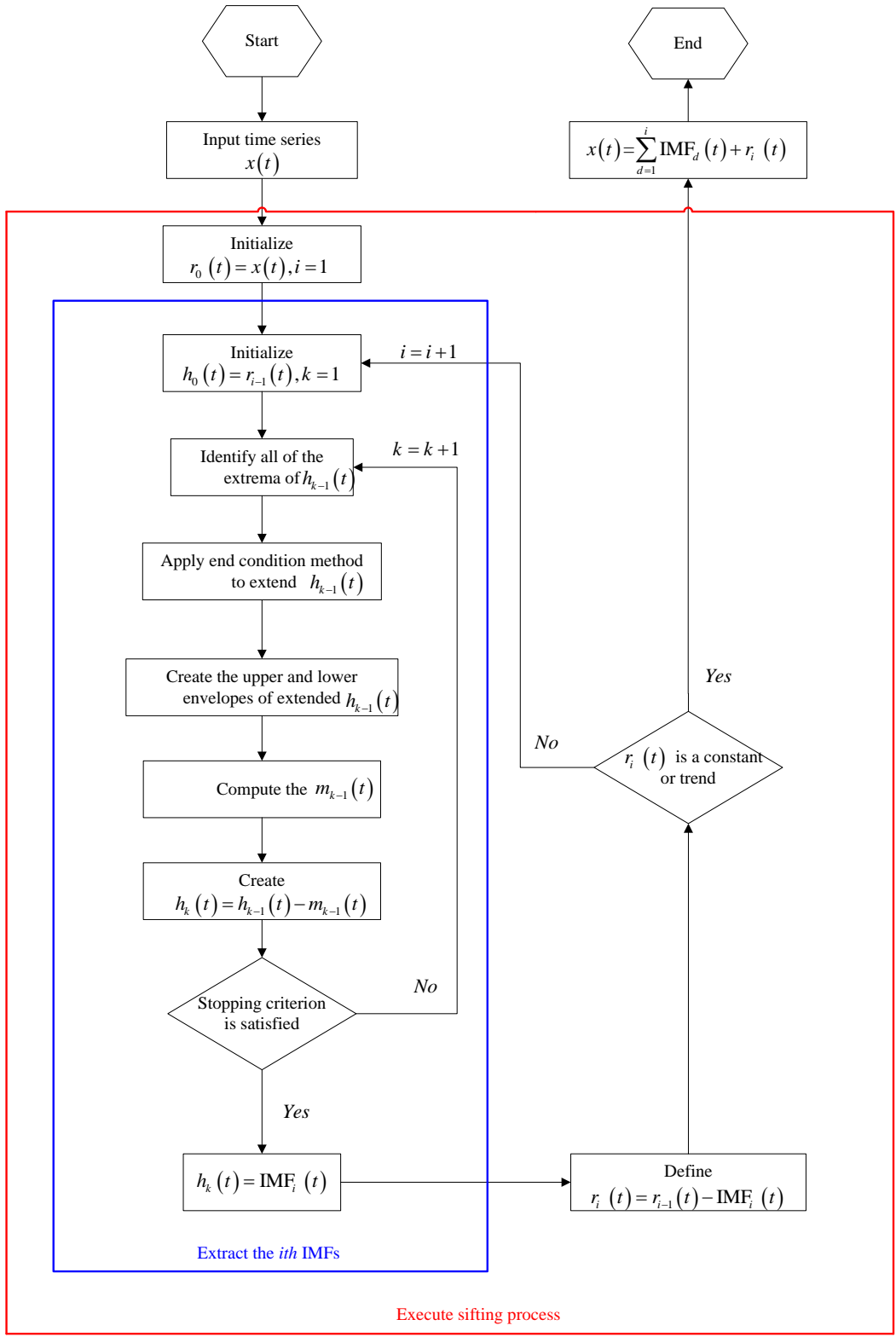

Fig. 1. The EMD with end condition method

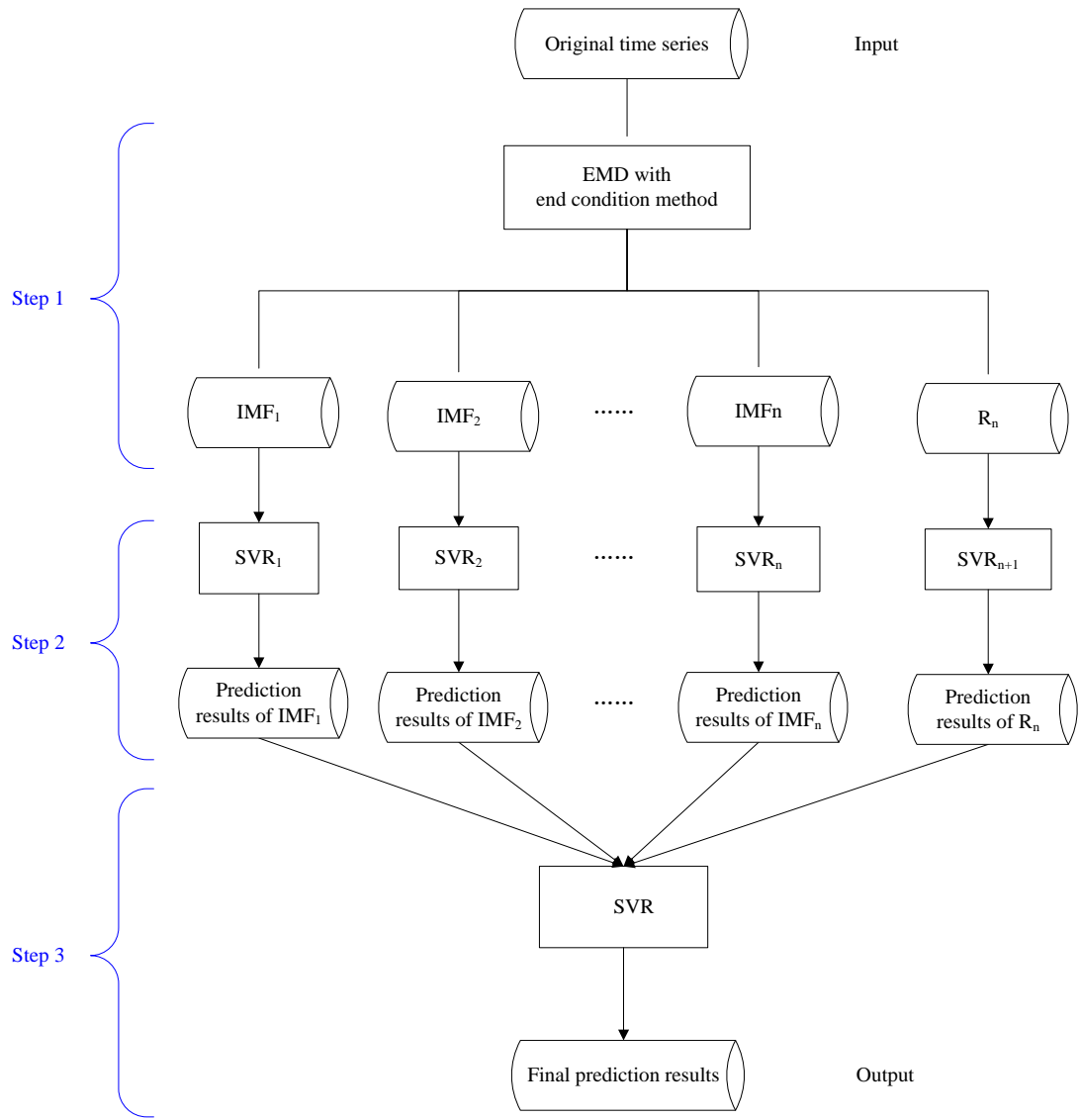

Fig. 2  The EMD-based SVR modeling framework

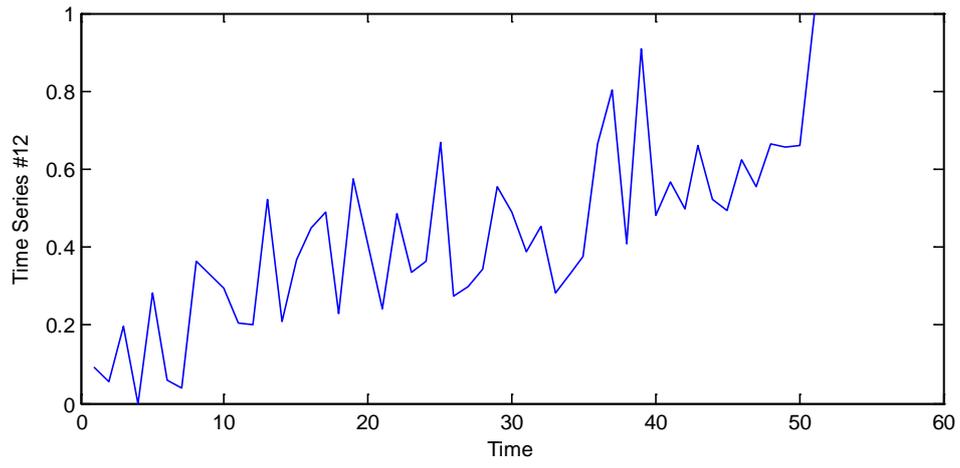
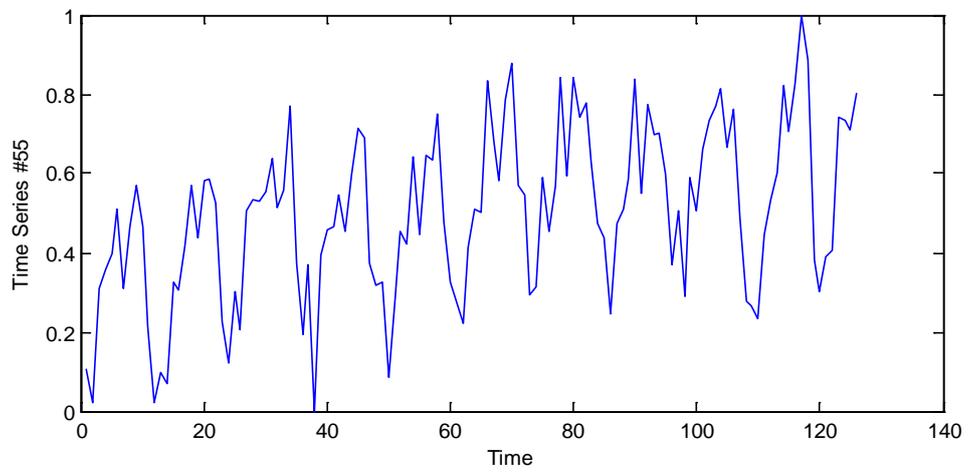
Fig. 3. Two representative NN3 time series (#12 and #55)

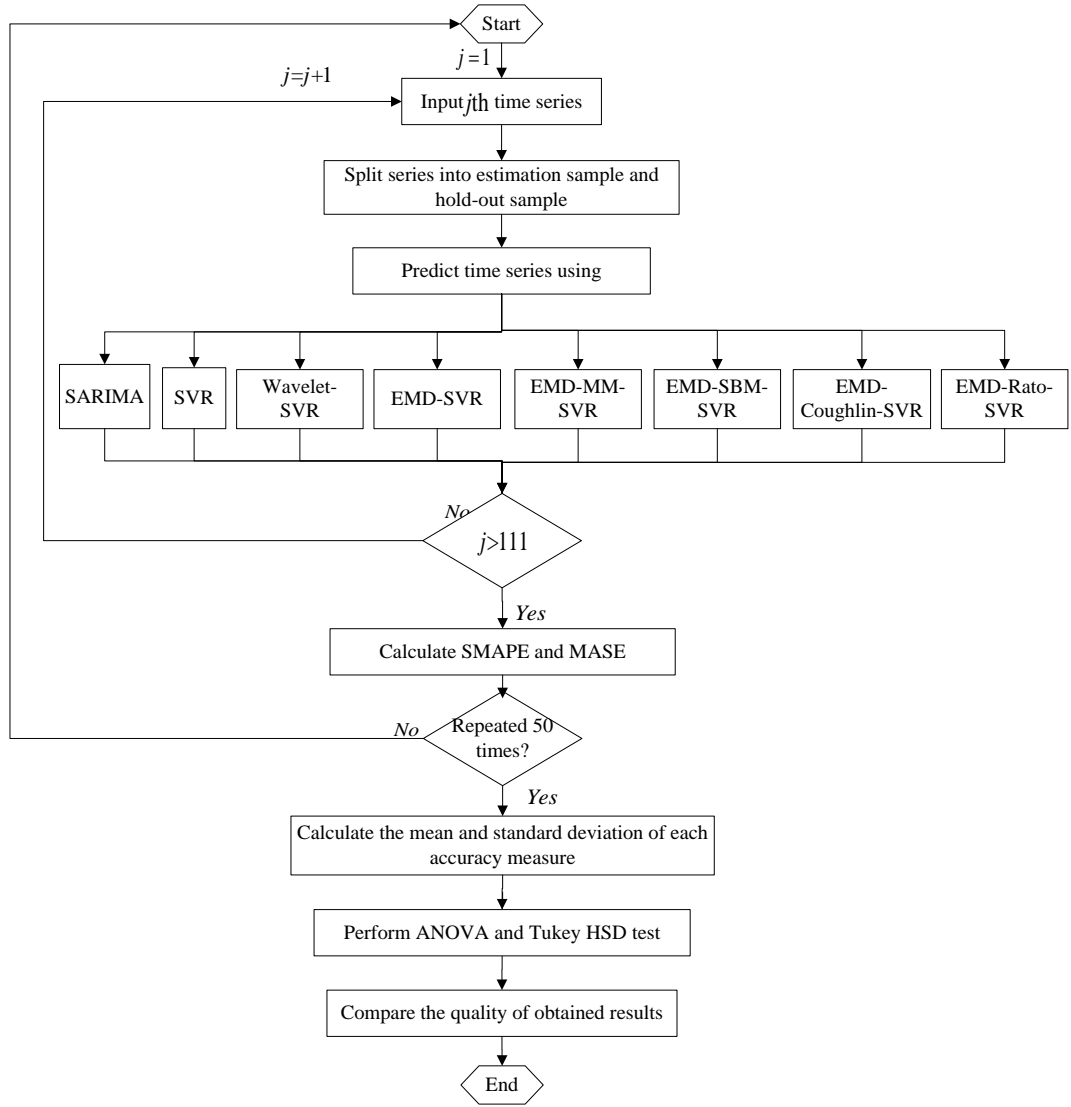

Fig. 4. Experiment procedures

# Tables

Table 1 Notation and definition for end condition methods

| Notation | Definition |
|---|---|
| $x(t)$ | The input time series, $x(t)=\{x(1),\ldots x(T)\}$ |
| $i$ | Index of local maximum, $i=1,\ldots,n$ |
| $j$ | Index of local minimum, $j=1,\ldots,m$ |
| $Max(1)$, $Min(1)$ | First two local extrema of time series $x(t)$ |
| $Max(n)$, $Min(m)$ | Last two local extrema of time series $x(t)$ |
| $P(i)$ | Ordinate value of $Max(i)$ respectively |
| $Q(j)$ | Ordinate value of $Min(j)$ respectively |
| $t(x(t))$ | Time index of $x(t)$ |
| $t(Max(i))$ | Time index of $Max(i)$ |
| $t(Min(j))$ | Time index of $Min(j)$ |
| $\Delta t_{max}(i)$ | Time gaps between two successive local maxima, $\Delta t_{max}(i)=t(Max(i+1))-t(Max(i))$ |
| $\Delta t_{min}(j)$ | Time gaps between two successive local minima, $\Delta t_{min}(j)=t(Min(j+1))-t(Min(j))$ |

Table 2 Prediction accuracy measure of different models for hold-out sample

| Prediction horizon | Model | SMAPE | | MASE | |
|---|---|---|---|---|---|
| | | Mean | Std | Mean | Std |
| $H=1$ | EMD-Rato-SVR | 7.854 | 0.0264 | 0.887 | 0.0026 |
| | EMD-Coughlin-SVR | 10.845 | 0.0315 | 0.956 | 0.0031 |
| | EMD-SBM-SVR | 6.494 | 0.0305 | 0.834 | 0.0029 |
| | EMD-MM-SVR | 11.084 | 0.0278 | 0.945 | 0.0025 |
| | EMD-SVR | 11.201 | 0.0295 | 1.006 | 0.0041 |
| | Wavelet-SVR | 12.012 | 0.0306 | 0.984 | 0.0037 |
| | SVR | 13.854 | 0.0297 | 1.113 | 0.0057 |
| | SARIMA | 17.125 | 0.0001 | 1.231 | 0.0000 |
| $H=18$ | EMD-Rato-SVR | 16.274 | 0.0321 | 1.187 | 0.0034 |
| | EMD-Coughlin-SVR | 18.005 | 0.0307 | 1.214 | 0.0065 |
| | EMD-SBM-SVR | 16.094 | 0.0285 | 1.196 | 0.0048 |
| | EMD-MM-SVR | 18.264 | 0.0312 | 1.424 | 0.0032 |
| | EMD-SVR | 20.241 | 0.0348 | 1.580 | 0.0047 |
| | Wavelet-SVR | 19.594 | 0.0371 | 1.612 | 0.0052 |
| | SVR | 22.254 | 0.0315 | 1.802 | 0.0038 |
| | SARIMA | 24.854 | 0.0001 | 2.216 | 0.0000 |

Table 3 ANOVA results for hold-out ample

| Prediction horizon | Measure | ANOVA Test | |
| --- | --- | --- | --- |
| | | Statistics F | p-value |
| $H=1$ | SMAPE | 29.815 | 0.000* |
| | MASE | 18.497 | 0.000* |
| $H=18$ | SMAPE | 9.640 | 0.001* |
| | MASE | 25.874 | 0.000* |

*Notes:** indicates the mean difference among the eight models is significant at the 0.05 level.

Table 4 Tukey HSD test results with ranked models for hold-out sample

| Prediction horizon | Measure | Ranks of models | | | | | | | | | | | | | |
|---|---|---|---|---|---|---|---|---|---|---|---|---|---|---|---|
| | | 1 | | 2 | | 3 | | 4 | | 5 | | 6 | | 7 | 8 |
| $H=1$ | SMAPE | SBM | <* | Rato | <* | Coughlin | < | MM | < | EMD-SVR | < | Wavelet-SVR | <* | SVR <* | SARIMA |
| | MASE | SBM | < | Rato | < | MM | < | Coughlin | < | Wavelet-SVR | < | EMD-SVR | <* | SVR <* | SARIMA |
| $H=18$ | SMAPE | SBM | < | Rato | <* | Coughlin | < | MM | <* | Wavelet-SVR | < | EMD-SVR | <* | SVR <* | SARIMA |
| | MASE | Rato | < | SBM | < | Coughlin | <* | MM | <* | EMD-SVR | < | Wavelet-SVR | <* | SVR <* | SARIMA |

*Notes:* **\*** indicates the mean difference between the two adjacent models is significant at the 0.05 level. 'SBM' corresponds to the EMD-SBM-SVR model, 'Rato' corresponds to the EMD-Rato-SVR model, 'Coughlin' corresponds to the EMD-Coughlin-SVR model, and 'MM' corresponds to the EMD-MM-SVR model.